\documentclass{edm_article}

\usepackage{framed}
\usepackage{graphicx}
\usepackage{subcaption}
\usepackage{color}
\usepackage{booktabs}
\usepackage{url}

\begin{document}

\title{Distinguishing Artificial from Authentic: Evaluating LLMs for Detecting LLM-Generated Content}

\numberofauthors{2}


\author{
Juho Leinonen \\
Aalto University \\
Espoo, Finland \\
juho.2.leinonen@aalto.fi
\and
Paul Denny \\
University of Auckland \\
Auckland, New Zealand \\
paul@cs.auckland.ac.nz
}

\maketitle

\begin{abstract}
As large language models (LLMs) are increasingly used by students to generate natural language responses and program code, there is growing interest in whether LLMs themselves can be used to distinguish AI-generated work from human-authored submissions. In this paper, we investigate the extent to which LLMs can detect their own generated content across multiple educational task types, including programming exercises, reflective writing, and short-answer questions. Using authentic student responses and multiple variants of LLM-generated answers, we evaluate detection performance under different prompting strategies and output formats. Our study addresses three research questions: (1) how accurately LLMs can identify their own outputs across task domains, (2) how detection effectiveness is influenced by factors such as prompt design, response length, and task type, and (3) what characteristics of LLM-generated responses contribute to successful or failed detection. Our findings show that LLM-based detection is highly task-dependent: detection is substantially more reliable for programming tasks and longer reflective responses, but performs poorly for short-answer questions, where LLMs frequently judge their own outputs as more human-like than authentic student responses. We further find that prompt framing and response verbosity have a pronounced effect on detectability in reflective writing tasks, with relatively minor prompt variations significantly reducing detection accuracy, while programming-related detection is more robust to prompt changes. Together, these results highlight both the potential and the limitations of LLM self-detection in educational settings and suggest caution in relying on LLMs as standalone tools for identifying AI-generated student work.
\end{abstract}


\section{Introduction}

Large language models (LLMs) are increasingly embedded in educational contexts.  Beyond their use by students to draft written responses \cite{jelson2024empirical} or generate code in programming courses \cite{kazemitabaar2024how}, LLMs are now being explored as graders, feedback providers, and generators of synthetic student data for research and tool development \cite{henkel2024can, menezes2024aigrading, leinonen2025llmitation}.  For synthetic data generation in particular, it is important for LLM-generated outputs to closely resemble authentic student work in both structure and distribution \cite{macneil2024synthetic}.  However, this goal exists alongside parallel efforts to reliably differentiate AI-generated outputs from authentic student work.

Interest in detection approaches for AI-generated content is growing in higher education contexts.  Universities have seen substantial growth in academic integrity referrals, with many thousands of cases linked to suspected AI use \cite{bergin2025acu}.  Concerns have also been raised around the reliability of detection approaches.  Some instructors have relied on direct LLM outputs to determine authorship, resulting in wrong accusations of misconduct \cite{ankel2023texas}.  Commonly used classifier-based detection tools, including systems such as GPTZero, have also been shown to be unstable and inconsistent \cite{malik2025ai}.  While watermarking approaches have been proposed, they are not widely accessible and they have been shown to be very fragile in some contexts, such as for code generation \cite{suresh2025watermarking}.  

Despite the growing interest, there has been relatively little research examining how well LLMs can detect LLM outputs in authentic educational task settings.  In particular, there has been limited investigation of how self-detection performance varies across domains such as programming, reflective writing, and short-answer responses, or how factors such as prompt framing and response length influence detectability. 
In this paper, we explore the following research questions:

\begin{itemize}
\item \textbf{RQ1}: How accurately can LLMs detect their own generated content across different educational task domains (programming, reflection, and short-answer responses)?
\item \textbf{RQ2}: How do prompt framing, response length, and task type influence LLM self-detection performance?
\item \textbf{RQ3}: What characteristics of LLM-generated responses contribute to successful or failed detection?
\end{itemize}

\section{Related Work}

As large language models (LLMs) are increasingly used to generate a wide range of content across many domains, there is growing interest in the ability to detect LLM-generated material \cite{wu2025survey}.  This is an essential problem for many reasons, including the prevention of misinformation and maintaining accountability in situations where human authorship is expected, such as in many educational contexts \cite{nathanson2024step, liu2024eduguard}.  There is also the problem of model collapse, as introduced by Shumailov et al. \cite{shumailov2024curserecursiontraininggenerated}, a phenomenon in which LLMs trained on data that includes their own generated content lose the ability to represent the original distribution of human-generated text.  As AI-generated text becomes more sophisticated, the difficulty of the detection task increases.  Traditional approaches tend to rely on innovations in watermarking techniques, statistical analysis detectors, and neural-network based detectors \cite{liu2024survey, abassy2024llmdetectaivetoolfinegrainedmachinegenerated, wu2025survey}. 

\subsection{LLM Self-Detection}

Recent work has also explored whether LLMs themselves can be effective detectors of LLM-generated content. For example, Zhu et al. propose a zero-shot black-box detection method that leverages ChatGPT to revise text and then measures the similarity between the original and revised versions \cite{zhu2023beat}. The intuition behind their approach is that ChatGPT will make fewer revisions to LLM-generated text compared to human-written text, as LLM outputs more closely align with the statistical patterns and generation logic learned by these models.  Similarly, Wang et al. apply instruction tuning of open-source LLMs to both document-level and sentence-level detection \cite{wang2024llmdetectorimprovingaigeneratedchinese}.  When applied to LLM-generated Chinese text, they find that instruction-tuned LLMs significantly outperform traditional classifiers.  Other recent work has shown that without modification, models such as GPT-4 and Llama 2 have non-trivial accuracy at distinguishing content they generate themselves from that of other LLMs and humans \cite{panickssery2024llmevaluatorsrecognizefavor}.
Steyvers et al. focus on the alignment between human trust in AI-generated responses and how LLMs communicate the likelihood of their predictions being correct \cite{Steyvers2025}.   Their findings suggest that improving uncertainty communication could enhance the reliability of LLM-generated content detection, particularly in educational and professional contexts.

\subsection{Challenges and Benchmarks}


Gressel et al. propose a benchmark for detecting AI-generated responses, testing both explicit and implicit detection strategies \cite{gressel2024humanadversarialbenchmarkexpose}. They find that explicit detection achieves a higher success rate, whereas implicit challenges, where LLMs are given a prompt to guide them to disguise their outputs, prove significantly more difficult.  Najjar et al. use explainable AI methods to classify text as human- or LLM-generated, outperforming some trained detection tools (like GPTZero) on both direct copying and more subtle forms of plagiarism \cite{najjar2025leveragingexplainableaillm}.

In the context of computing education, Hoq et al. present an ML-based approach to detect ChatGPT-generated code submissions in CS1 courses, achieving over 90\% accuracy with traditional ML and AST-based deep learning models \cite{hoq2024detecting}. Their findings reveal that even with prompt variations intended to mimic novice programmers, ChatGPT-generated code remains structurally distinct from student code. They observe that code structure provides detectable signatures that differentiate student work from AI-generated solutions, reinforcing the idea that LLMs could be trained to detect their own outputs in structured domains.  In contrast, Orenstrakh et al. evaluate eight LLM-generated text detectors in computing education, noting that detection accuracy decreases with code-based submissions and non-English text~\cite{orenstrakh2023detectingllmgeneratedtextcomputing}.

\subsection{Biases and Misinformation}

Other work has focused more explicitly on biases and detecting misinformation. 
Chen et al. explore the challenges of detecting LLM-generated misinformation, demonstrating that such content is often harder to detect than human-written misinformation due to its deceptive style \cite{chen2024llmgeneratedmisinformationdetected}.  They propose a taxonomy for LLM-generated misinformation and highlight the need for robust detection frameworks. Lin et al. investigate biases in LLM-based detection systems, revealing that political and linguistic biases can affect classification accuracy \cite{lin2024investigatingbiasllmbasedbias}. They propose debiasing techniques to improve fairness in AI-generated content detection.
Beyond simple detection, Achintalwar et al. focus on identifying risks within LLM-generated content, such as bias, hallucinations, toxicity, and misinformation, rather than distinguishing AI-generated text from human-written text \cite{achintalwar2024detectorssafereliablellms}. They introduce a library of compact classifiers that can operate independently of the underlying LLMs, allowing for scalable AI safety measures.

\section{Methods}

\begin{figure}[t]
\Description{You have been learning MATLAB for 6 weeks now, and have just begun to learn the C programming language. You will notice that some things are different, and some are similar. Perhaps you find it easier to learn another language, given that you already understand basic programming principles, or perhaps you find it harder because some things work differently to what you are now familiar with. Everyone will experience this differently.

For this exercise, write a short reflective piece (anywhere from several sentences to several paragraphs) in your own words, that describes how you have found this transition to learning a new language. What are you finding easier, and what are you finding harder about learning C compared to MATLAB?}
    \centering
    \begin{framed}
    \scriptsize
     \raggedright
        You have been learning MATLAB for 6 weeks now, and have just begun to learn the C programming language. You will notice that some things are different, and some are similar. Perhaps you find it easier to learn another language, given that you already understand basic programming principles, or perhaps you find it harder because some things work differently to what you are now familiar with. Everyone will experience this differently.

        For this exercise, write a short reflective piece (anywhere from several sentences to several paragraphs) in your own words, that describes how you have found this transition to learning a new language. What are you finding easier, and what are you finding harder about learning C compared to MATLAB?
    \end{framed}
    \caption{Reflection: an exercise eliciting students' reflections on transitioning from MATLAB to C in an introductory programming course.}
    \Description[Reflection prompt]{Reflection prompt}
    \label{fig:reflection_box}
\end{figure}

\begin{figure}[h]
\Description{A figure showing two code snippet examples where the examples are identical except for the spacing used. Students are asked: `Ìn a sentence, why do you think the code you selected above is easier to read?}
    \centering
    \begin{framed}
    \scriptsize
     \raggedright
        You have now seen lots of examples of code.  The two examples shown below are identical except for the spacing used.  
        
        \centering
        \includegraphics[width=\columnwidth]{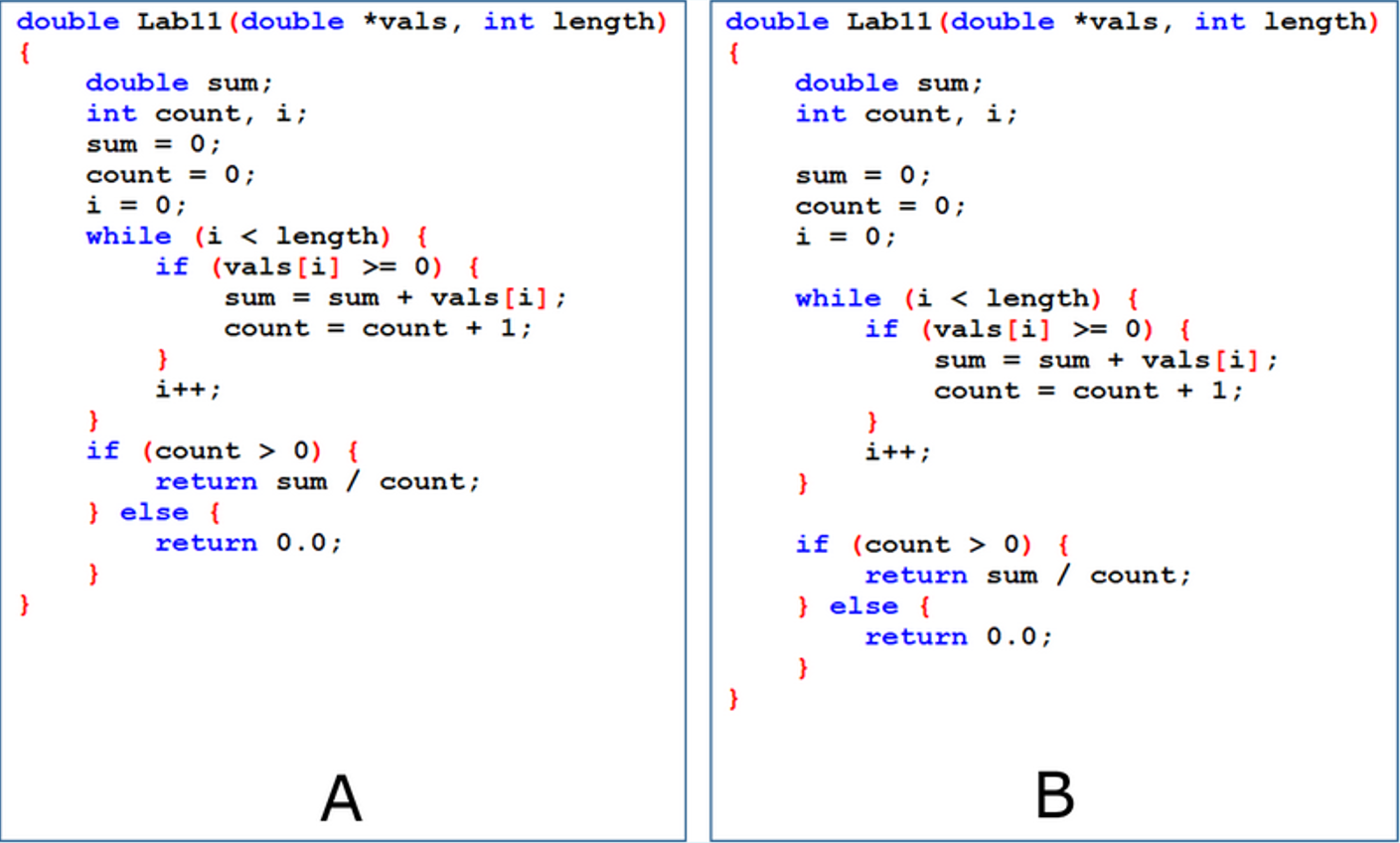} 
        \raggedright        

        In a sentence, why do you think the code you selected above is easier to read?
    \end{framed}
    \caption{Short answer: an exercise eliciting a sentence-length response from students on the readability of a provided code fragment.}
    \Description[Readability prompt]{Readability prompt}
    \label{fig:shortanswer_box}
\end{figure}

\begin{figure}[h]
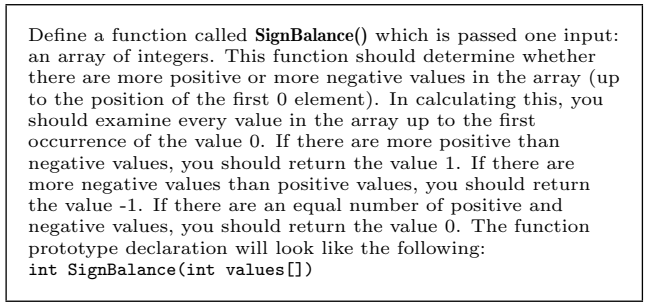

\Description{Define a function called \textbf{SignBalance()} which is passed one input: an array of integers. This function should determine whether there are more positive or more negative values in the array (up to the position of the first 0 element).  In calculating this, you should examine every value in the array up to the first occurrence of the value 0.  If there are more positive than negative values, you should return the value 1. If there are more negative values than positive values, you should return the value -1. If there are an equal number of positive and negative values, you should return the value 0. 
The function prototype declaration will look like the following:

    \texttt{int SignBalance(int values[])}}
    \centering
    \begin{framed}
    \scriptsize
     \raggedright
        Define a function called \textbf{SignBalance()} which is passed one input: an array of integers. This function should determine whether there are more positive or more negative values in the array (up to the position of the first 0 element).  In calculating this, you should examine every value in the array up to the first occurrence of the value 0.  If there are more positive than negative values, you should return the value 1. If there are more negative values than positive values, you should return the value -1. If there are an equal number of positive and negative values, you should return the value 0. 
The function prototype declaration will look like the following:

    \texttt{int SignBalance(int values[])}

    \end{framed}
    \caption{Programming 1: an exercise involving loops and arrays, where answers should be specified in the C language.}
    \Description[Programming prompt 1]{Programming prompt 1}
    \label{fig:prog1_box}
\end{figure}

\begin{figure}[h]
\Description{Write a function that takes two strings as input: the word to check and the pattern against which to check it.  The function prototype declaration is as follows:

    \texttt{int WordMatchesPattern(char *word, char *pattern)}

The function should return true only if the string ``word'' is a possible completion of the string ``pattern''.  The ``pattern'' string is a partial word where some of the characters have been replaced with hyphens (-).  The “word” will be a valid completion of the ``pattern'' if all of the characters not corresponding to hyphens match, and if they are both the same length. }
    \centering
    \begin{framed}
    \scriptsize
     \raggedright
Write a function that takes two strings as input: the word to check and the pattern against which to check it.  The function prototype declaration is as follows:

    \texttt{int WordMatchesPattern(char *word, char *pattern)}

The function should return true only if the string ``word'' is a possible completion of the string ``pattern''.  The ``pattern'' string is a partial word where some of the characters have been replaced with hyphens (-).  The “word” will be a valid completion of the ``pattern'' if all of the characters not corresponding to hyphens match, and if they are both the same length. 

    \end{framed}
    \caption{Programming 2: an exercise involving string matching, where answers should be specified in the C language.}
    \Description[Programming prompt 2]{Programming prompt 2}
    \label{fig:prog2_box}
\end{figure}

\subsection{Context and Data}

We analyse student submissions from a university-level introductory programming course.  The course was taught over 12 weeks at the University of Auckland, and is a core (compulsory) course that engineering students take in their first year of study.  The first half of the course uses the MATLAB language, and from week 7 students transition to the C language.   

Students completed weekly lab sessions, and our data was gathered from labs delivered during weeks 7, 9, 10 and 11, across four different exercises.  The week 7 data are responses to a reflection task (\emph{reflection}) shown in Figure \ref{fig:reflection_box}, asking students about their perceptions of the switch to a new language.  Responses were expected to be several sentences to several paragraphs.  The Week 9 and 10 data are responses to two programming exercises: the first (\emph{programming 1}) to determine whether there are more positive or more negative values in the array (up to the position of the first 0 element), and the second (\emph{programming 2}) is a string matching exercise to determine if a string could be a valid completion of a partial string with missing letters.  The tasks are shown in Figures \ref{fig:prog1_box} and \ref{fig:prog2_box}.  Finally, the Week 11 data is a short answer task (\emph{short answer}) on the readability of code, where expected responses were a single sentence, as shown in Figure \ref{fig:shortanswer_box}. 

The number of submissions per exercise was: programming 1 ($n=201$), programming 2 ($n=870$), short answer ($n=854$), and reflection ($n=846$). In total, the dataset includes submissions from 913 unique students across all exercises.

\begin{table*}[h]
    \centering
    \caption{Study Results for Reflection, Short Answer, and Programming Questions}
    \label{tab:study_results}
    \begin{tabular}{lccccccccc}
        \toprule
        Source & \multicolumn{2}{c}{Likelihood (\%)} & Yes (\%) & No (\%) & \multicolumn{2}{c}{Length} & \multicolumn{3}{c}{Differences} \\
        \cmidrule(lr){2-3} \cmidrule(lr){6-7} \cmidrule(lr){8-10}
        & Mean & Std Dev & & & Mean & Std Dev & Likelihood & Length & Yes (\%) \\
        \midrule
        \multicolumn{10}{c}{\textbf{Reflection Question}} \\
        Real Students & 68.8  & 25.2 & 16.8 & 83.2 & 439.7  & 316.6 & -     & -     & - \\
        Baseline LLM  & 87.0  & 7.9  & 63.8 & 36.2 & 2329.4 & 197.0 & 18.2  & 1889.7& 47.0 \\
        Persona LLM   & 81.2  & 16.4 & 20.6 & 79.4 & 1892.6 & 209.9 & 12.4  & 1452.9& 3.8 \\
        Typo LLM      & 74.8  & 22.1 & 6.2  & 93.8 & 1577.9 & 180.4 & 6.1   & 1138.2& -10.5 \\
        \midrule
        \multicolumn{10}{c}{\textbf{Short Answer Question}} \\
        Real Students & 74.1  & 23.4 & 66.7 & 33.3 & 91.6   & 65.7  & -     & -     & - \\
        Baseline LLM  & 68.6  & 26.7 & 30.4 & 69.6 & 214.7  & 31.9  & -5.5  & 123.1 & -36.3 \\
        Persona LLM   & 72.6  & 20.6 & 3.4  & 96.6 & 203.7  & 30.8  & -1.5  & 112.1 & -63.3 \\
        Typo LLM      & 51.7  & 32.2 & 0.4  & 99.6 & 305.7  & 44.7  & -22.4 & 214.1 & -66.3 \\
        \midrule
        \multicolumn{10}{c}{\textbf{Programming Question 1}} \\
        Real Students & 59.5  & 41.4 & 1.2  & 98.8 & 390.2  & 112.1 & -     & -     & - \\
        Baseline LLM  & 80.0  & 29.2 & 5.3  & 94.7 & 663.5  & 113.6 & 20.5  & 273.3 & 4.1 \\
        Persona LLM   & 80.2  & 29.0 & 5.1  & 94.9 & 653.6  & 127.7 & 20.7  & 263.5 & 3.9 \\
        Bugs LLM      & 86.1  & 18.8 & 9.7  & 90.3 & 853.3  & 157.2 & 26.6  & 463.2 & 8.6 \\
        \midrule
        \multicolumn{10}{c}{\textbf{Programming Question 2}} \\
        Real Students & 73.1  & 33.4 & 2.8  & 97.3 & 405.7  & 173.7 & -     & -     & - \\
        Baseline LLM  & 91.0  & 9.4  & 42.1 & 57.9 & 676.5  & 83.5  & 17.9  & 270.8 & 39.4 \\
        Persona LLM   & 90.7  & 9.6  & 42.1 & 57.9 & 622.6  & 111.6 & 17.7  & 216.9 & 39.3 \\
        Bugs LLM      & 88.5  & 12.5 & 35.3 & 64.7 & 736.5  & 155.9 & 15.4  & 330.8 & 32.6 \\
        \bottomrule
    \end{tabular}
\end{table*}

\begin{table}[t]
\centering
\small
\caption{Rank-based separability of detector likelihood scores between real student responses and LLM-generated responses. We report Cliff's $\delta$ effect sizes and Holm-adjusted $p$-values from two-sided Mann--Whitney U tests.}
\label{tab:separability}
\begin{tabular}{llrrll}
\toprule
Exercise & Condition & $n_{\text{real}}$ & $n_{\text{cond}}$ & Cliff's $\delta$ & $p_{\mathrm{Holm}}$ \\
\midrule
Refl.   & Baseline & 846 & 100 & -0.841 & 1.28e-42 \\
Refl.   & Persona  & 846 & 100 & -0.451 & 3.26e-13 \\
Refl.   & Typo     & 846 & 100 & -0.131 & 0.0322 \\
\midrule
Short ans.  & Baseline & 854 & 100 &  0.162 & 0.0078 \\
Short ans.  & Persona  & 854 & 100 &  0.286 & 5.64e-06 \\
Short ans.  & Typo     & 854 & 100 &  0.608 & 7.17e-23 \\
\midrule
Prog. 1 & Baseline & 201 & 100 & -0.473 & 4.47e-11 \\
Prog. 1 & Persona  & 201 & 100 & -0.472 & 4.47e-11 \\
Prog. 1 & Bugs     & 201 & 100 & -0.640 & 4.64e-19 \\
\midrule
Prog. 2 & Baseline & 870 & 100 & -0.699 & 6.08e-30 \\
Prog. 2 & Persona  & 870 & 100 & -0.670 & 9.2e-28 \\
Prog. 2 & Bugs     & 870 & 100 & -0.512 & 4.77e-17 \\
\bottomrule
\end{tabular}
\end{table}

\begin{figure*}[t]
\Description{Four density plots of LLM-likelihood (0-100\%) comparing real students vs 3 LLM variants; reflection/programming higher for LLM; short-answer reversed.}

    \centering

    \begin{subfigure}[t]{0.49\linewidth}
        \centering
        \includegraphics[width=\linewidth]{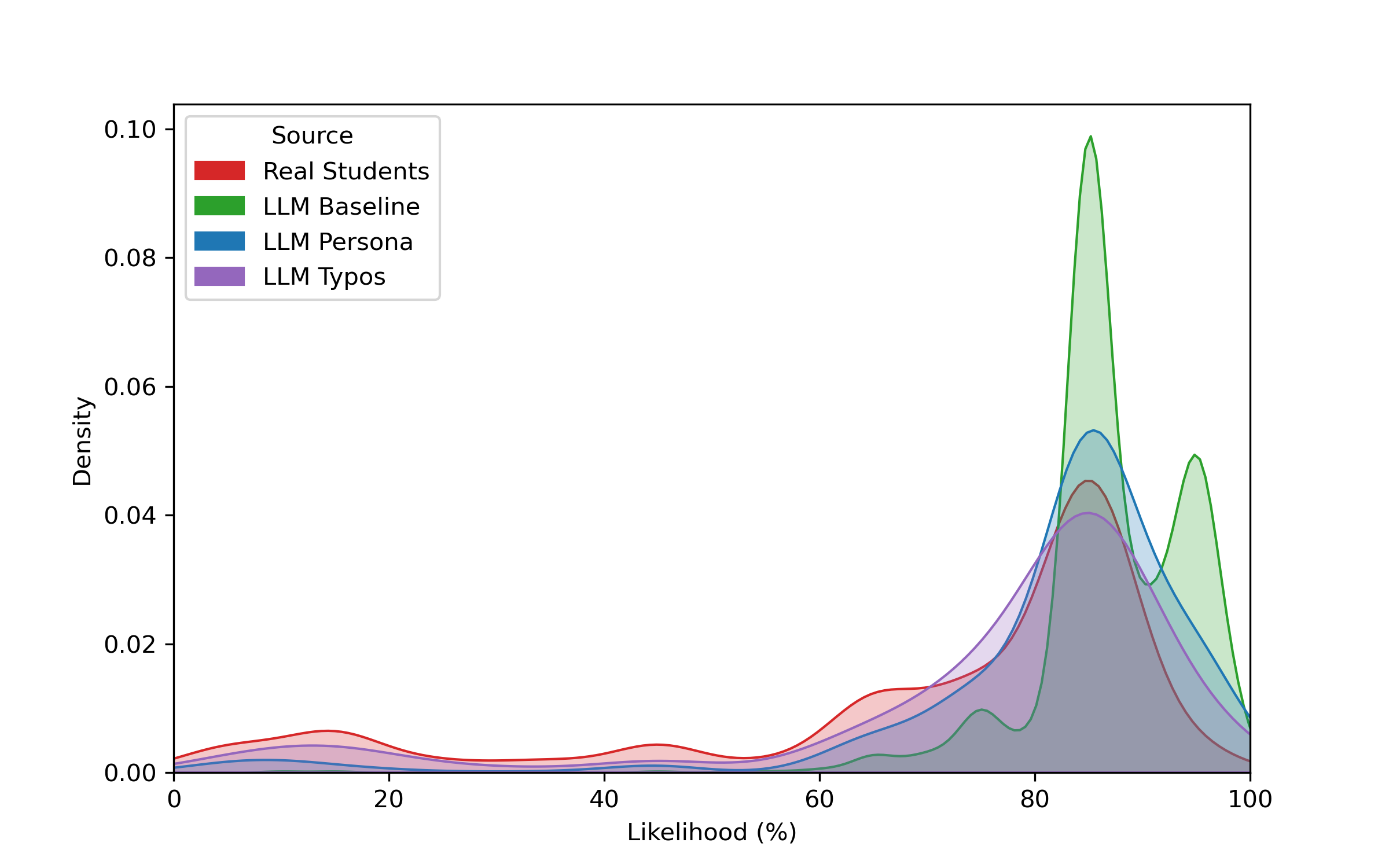}
        \caption{Reflection}
        \label{fig:dist_reflection}
    \end{subfigure}
    \hfill
    \begin{subfigure}[t]{0.49\linewidth}
        \centering
        \includegraphics[width=\linewidth]{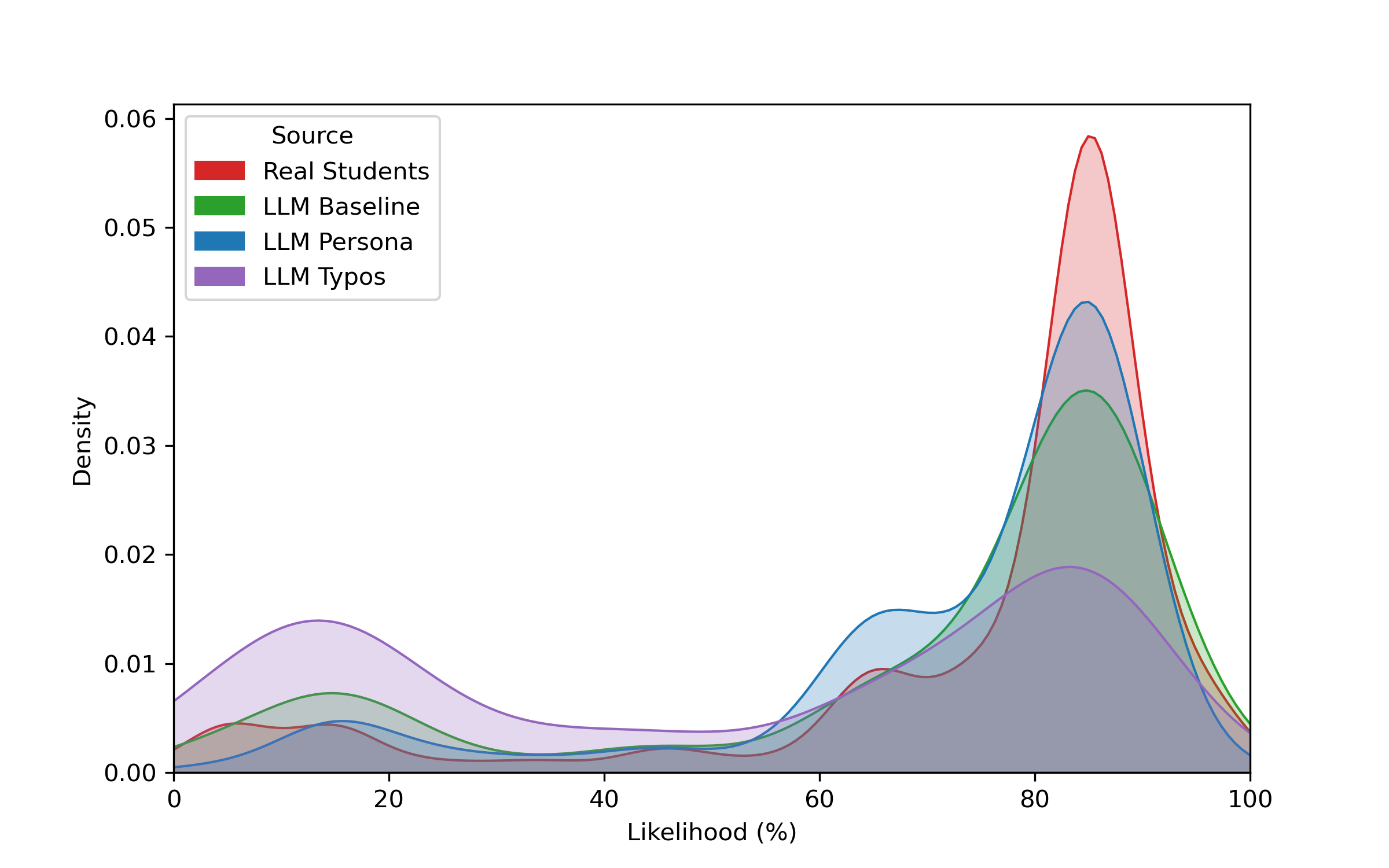}
        \caption{Short answer}
        \label{fig:dist_shortanswer}
    \end{subfigure}


    \begin{subfigure}[t]{0.49\linewidth}
        \centering
        \includegraphics[width=\linewidth]{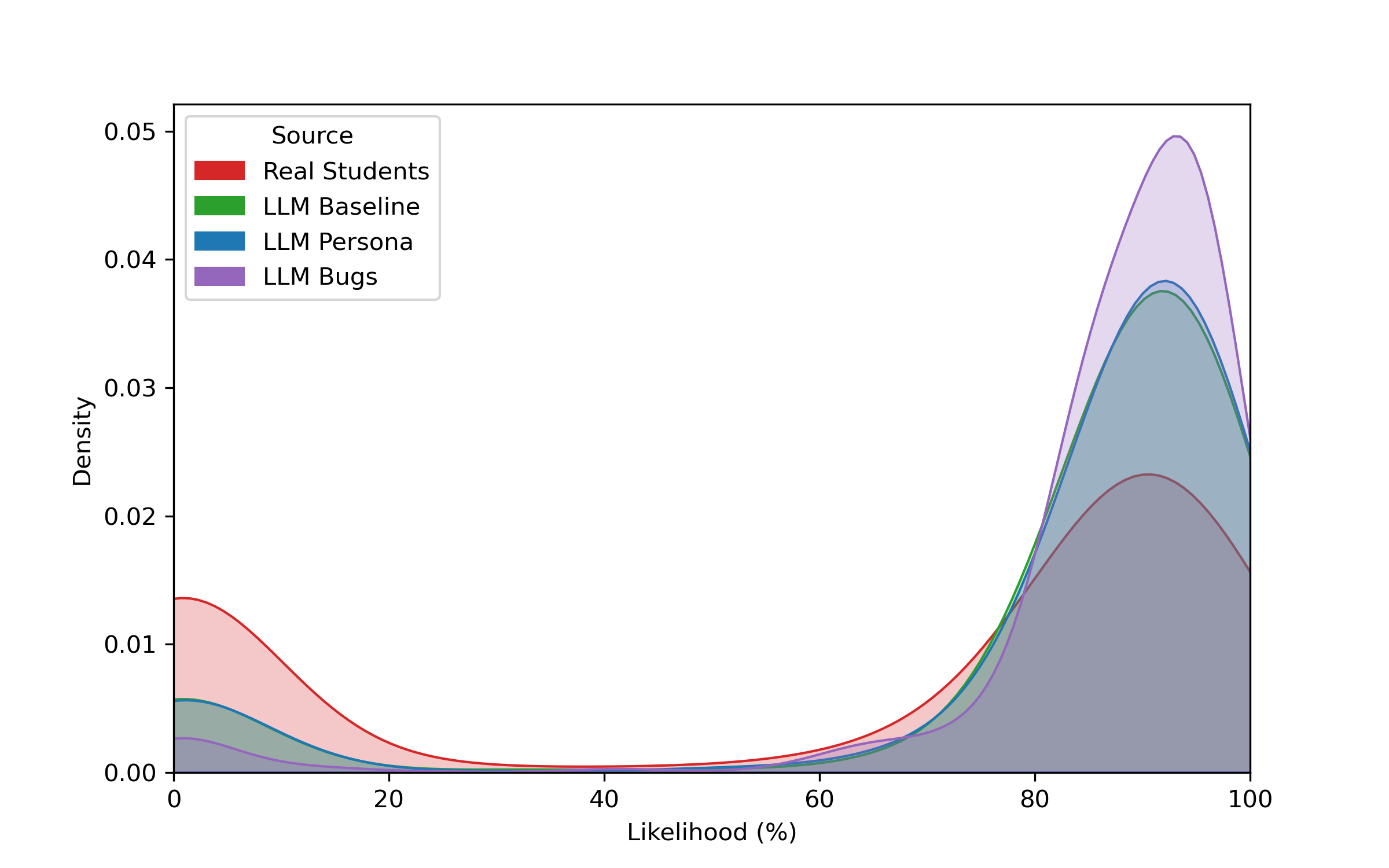}
        \caption{Programming 1}
        \label{fig:dist_programming1}
    \end{subfigure}
    \hfill
    \begin{subfigure}[t]{0.49\linewidth}
        \centering
        \includegraphics[width=\linewidth]{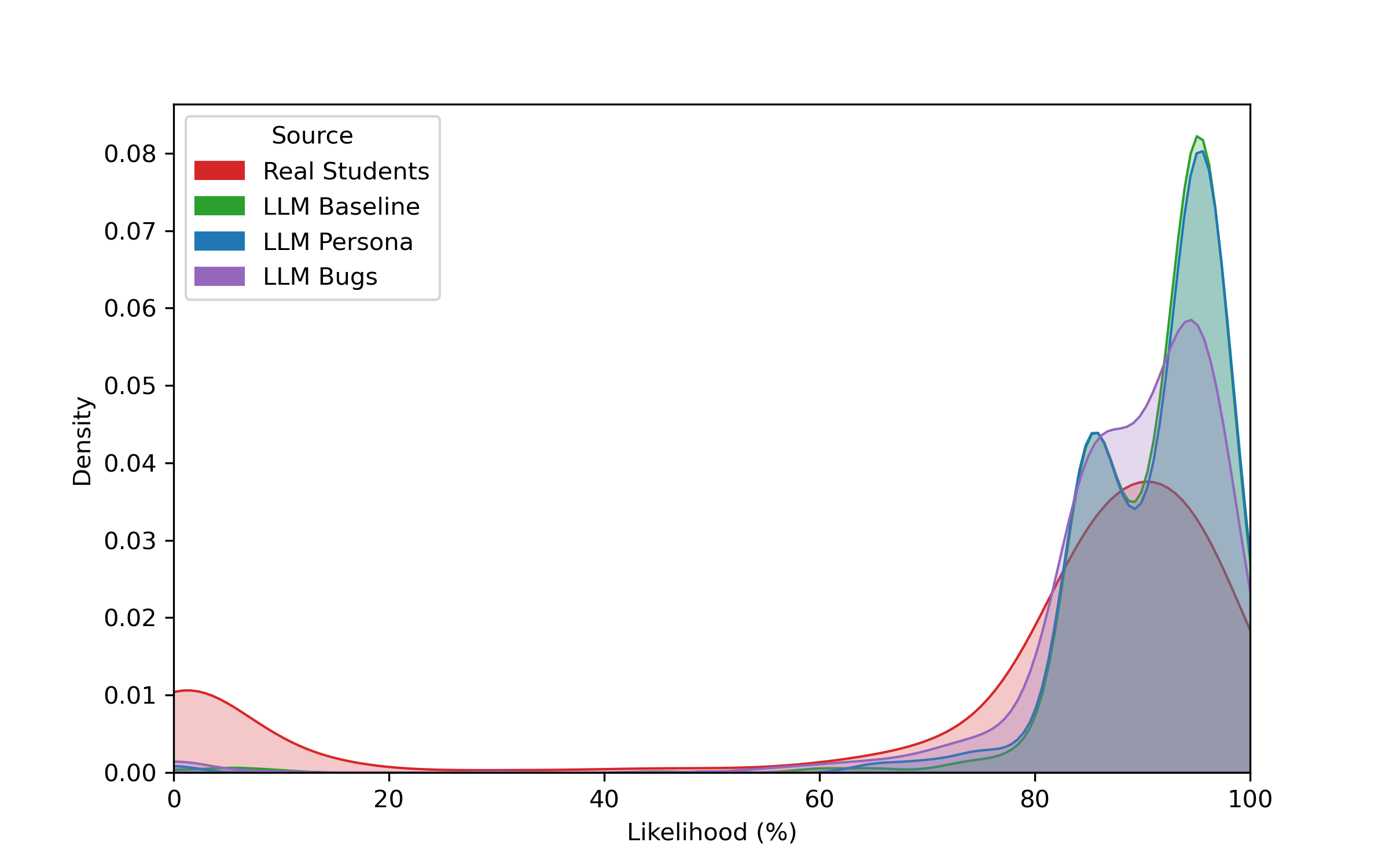}
        \caption{Programming 2}
        \label{fig:dist_programming2}
    \end{subfigure}

    \caption{Distributions for each question type for the likelihood prompt.}
    \label{fig:all_distributions}
\end{figure*}

\subsection{Generating LLM Data}
We generated synthetic responses using GPT-4o under three prompting conditions per exercise type. The prompts included the content of the prompts that were shown to students as part of the labs tasks (see Figures 1--4). 

For the textual exercises (reflection and short answer), we used:
(i) \emph{baseline}: ``Your job is to generate responses to exercises.''
(ii) \emph{persona}: ``Your job is to generate responses to exercises. Make it seem like a student response in an introductory programming course.''
(iii) \emph{typos}: ``Your job is to generate responses to exercises. Make it seem like a student response in an introductory programming course. Make sure to include a lot of typos and use informal language.''

For the programming exercises (programming 1 and programming 2), we used:
(i) \emph{baseline}: ``Your job is to generate responses to exercises.''
(ii) \emph{persona}: ``Your job is to generate responses to exercises. Make it seem like a student response in an introductory programming course.''
(iii) \emph{bugs}: ``Your job is to generate responses to exercises. Make it seem like a student response in an introductory programming course. The response should include common beginner mistakes and exhibit poor programming style.''

For each prompt--exercise combination, we generated 100 responses, yielding
$2 \text{ textual exercises} \times 3 \text{ prompts} \times 100 = 600$ textual LLM responses and
$2 \text{ programming exercises} \times 3 \text{ prompts} \times 100 = 600$ programming LLM responses,
for a total of 1200 LLM-generated responses.

\subsection{Evaluation}
We evaluate detectability by prompting an LLM to judge whether a response was LLM-generated. For the programming exercises, we removed non-code text from LLM-generated responses to match the student submission format (students submitted code only).

We used two detector prompts:
(i) \emph{likelihood}: ``Your job is to detect LLM-generated responses. Only ever respond with a percentage indicating how likely it is that the response was generated by an LLM.''
(ii) \emph{yes/no}: ``Your job is to detect LLM-generated responses. Only ever respond with yes or no indicating whether you think that the response was generated by an LLM (yes = it was generated by an LLM, no = it was not generated by an LLM).''

To account for variability in the detector's outputs, we ran the detector 10 times per response for both prompts.

\subsection{Analysis}
We report descriptive statistics for response length (mean and standard deviation) and detector outputs. For the likelihood prompt, we report the mean and standard deviation of likelihood values pooled across all detector runs (10 runs per response). For the yes/no prompt, we report the percentage of ``yes'' and ``no'' outputs pooled across all runs.

To quantify separability between real student responses and LLM-generated responses, we compare detector likelihood scores between (i) real student responses and (ii) each LLM prompting condition, separately for each exercise. Because likelihood scores are obtained from repeated detector runs, inferential analyses are performed at the response level: we first average the 10 likelihood outputs for each response so that each response contributes a single score. We then use Mann--Whitney U tests (two-sided) and report Cliff's delta ($\delta$) as an effect size. We adjust $p$-values for multiple comparisons using Holm's method~\cite{holm1979simple}.

\section{Results}

Table~\ref{tab:study_results} summarizes descriptive statistics for detector outputs and response lengths, while Table~\ref{tab:separability} quantifies rank-based separability of the likelihood scores between real student responses and each LLM prompting condition. Figure~\ref{fig:all_distributions} visualizes the distributions of likelihood estimates.

\subsection{Overall Patterns Across Tasks}
Across three of the four exercises (reflection, programming 1, programming 2), LLM-generated responses generally received higher likelihood estimates than real student responses (Table~\ref{tab:study_results}), and the likelihood distributions were shifted toward higher values for LLM responses (Figure~\ref{fig:dist_reflection}, Figure~\ref{fig:dist_programming1}, Figure~\ref{fig:dist_programming2}). This is also reflected in predominantly negative Cliff's $\delta$ values in Table~\ref{tab:separability} (computed as $\delta(\text{real}, \text{LLM})$, so negative values indicate higher likelihoods for LLM responses).

In contrast, the short answer exercise exhibited the opposite pattern: real student responses tended to receive \emph{higher} likelihood estimates than LLM-generated responses (Table~\ref{tab:study_results}), with positive Cliff's $\delta$ across all three prompting conditions (Table~\ref{tab:separability}). In Figure~\ref{fig:dist_shortanswer}, the real-student distribution is centered slightly to the right of the baseline and persona distributions, and the typo condition shows substantial probability mass at low likelihood values.

\subsection{Reflection and Short Answer (Textual Exercises)}

\begin{table*}[t]
\centering
\scriptsize
\caption{Example responses for the short answer question by source.}
\label{tab:text-example-responses}
\begin{tabular}{p{0.27\columnwidth} p{1.7\columnwidth}}
\toprule
\textbf{Source} & \textbf{Example} \\
\midrule
Real Students &
the appropriate spacing between parts of the function makes B easier to read. \\ [3pt]

Baseline LLM &
Code B is easier to read because the additional line breaks help to visually separate different sections of the code, such as variable declarations, initializations, the loop, and the conditional statement, making it easier to follow the logic structure. \\ [3pt]

Persona LLM &
In my opinion, Code B is easier to read because the additional spacing and line breaks help separate the different logical sections of the code, making it easier to follow the flow and understand the structure. \\ [3pt]

Typo LLM &
I think Code B is easier to read becoz it has more spaces and indentations. It makes it look cleaner and its easier to spot where different parts of the code are, like the initialization and the loop and if statements. It's like when u write an essay and use paragraphs, everything isn't all squished together.  \\

\bottomrule
\end{tabular}
\end{table*}

For the reflection question, baseline LLM responses were substantially more detectable than real student responses by the likelihood prompt (means 87.0 vs.\ 68.8; Table~\ref{tab:study_results}). The separation was large and statistically reliable (Cliff's $\delta=-0.841$, $p_{\mathrm{Holm}}=1.28\mathrm{e}{-42}$; Table~\ref{tab:separability}). Figure~\ref{fig:dist_reflection} shows a pronounced concentration of likelihoods for baseline LLM near the upper range, whereas real student responses exhibit a broader distribution with more mass at lower likelihoods. Persona prompting reduced separability relative to baseline (Cliff's $\delta=-0.451$), and the typo prompt reduced separability further (Cliff's $\delta=-0.131$), with the typo distribution overlapping more closely with real students (Figure~\ref{fig:dist_reflection}).

For the short answer question, the detector behaved differently. Real student responses had a higher mean likelihood than any LLM condition (74.1 for real vs.\ 68.6 baseline, 72.6 persona, and 51.7 typo; Table~\ref{tab:study_results}). Rank-based comparisons confirm this reversal (Cliff's $\delta=0.162$ for baseline, $0.286$ for persona, and $0.608$ for typo; all $p_{\mathrm{Holm}} \le 0.0078$; Table~\ref{tab:separability}). Figure~\ref{fig:dist_shortanswer} suggests that the typo condition, in particular, often received low likelihood estimates, indicating that the detector frequently judged these LLM responses as human-like despite being generated.

Table \ref{tab:text-example-responses} shows one example from each source of a response to the short answer question.  Responses from real students tended to be shorter, as in the example, whereas the typo LLM included words such as ``becoz'' and ``u''.

\subsection{Programming Exercises}
For programming question 1, LLM responses received substantially higher likelihood estimates than real student code (means 80.0--86.1 for LLM vs.\ 59.5 for real; Table~\ref{tab:study_results}). The likelihood distributions in Figure~\ref{fig:dist_programming1} show real student responses with considerable mass near very low likelihood values, while LLM responses are concentrated near high likelihoods. Rank-based separability was moderate to large (Cliff's $\delta\approx-0.47$ for baseline/persona and $\delta=-0.640$ for bugs; all $p_{\mathrm{Holm}}<10^{-10}$; Table~\ref{tab:separability}), suggesting that adding beginner mistakes and poor style did not reduce detectability for this question and may have increased it.

For programming question 2, likelihood estimates were higher overall and separation was stronger for baseline and persona prompting (means 91.0 and 90.7 vs.\ 73.1 for real; Table~\ref{tab:study_results}), with large negative effect sizes (Cliff's $\delta=-0.699$ baseline and $-0.670$ persona; Table~\ref{tab:separability}). However, the bugs prompt reduced separability relative to baseline/persona (Cliff's $\delta=-0.512$), which is also visible in Figure~\ref{fig:dist_programming2} as increased overlap between the bugs and real-student distributions compared to baseline/persona.

\subsection{Response Length as a Potential Confound}
Across both textual exercises, LLM-generated responses were substantially longer than real student responses (Table~\ref{tab:study_results}). For reflection, LLM responses were roughly 3.5--5$\times$ longer on average (1577.9--2329.4 vs.\ 439.7 characters). For short answers, LLM responses were about 2--3$\times$ longer on average (203.7--305.7 vs.\ 91.6). For programming exercises, LLM code was also longer than student code (e.g., 663.5--853.3 vs.\ 390.2 characters in programming question 1). These consistent length differences suggest that response length may be one signal the detector can exploit, particularly for the text-based exercises.

\section{Discussion}


\subsection{Task Differences}
For RQ1, \textit{How accurately can LLMs detect their own generated content across different educational task domains?}, 
self-detection performance was strongly \textit{task-dependent}. Detection was most reliable for the two programming tasks and for the longer reflective responses, where LLM-generated submissions typically received higher ``LLM-likelihood'' scores than real student work. In contrast, detection performed poorly for the short-answer task: it was the only task type where both the likelihood estimates and binary judgments were higher for real student responses than for LLM-generated ones, indicating a clear failure mode where brief, one-sentence answers were not only hard to classify but could be systematically misranked. This suggests that conclusions about detectability are unlikely to transfer uniformly across assessment formats, even within the same course context.

\subsection{Prompting and Length}
For RQ2, \textit{How do prompt framing, response length, and task type influence LLM self-detection performance?}, 
both detector prompting and generation prompting influenced outcomes, and their effects varied by task type. On the detector side, asking for a graded likelihood estimate produced more consistent separation than forcing a binary yes/no decision. For LLM-generated content, mean likelihood estimates were always in the correct direction (frequently high), whereas binary decisions tended to misclassify LLM-generated responses across most task and prompt combinations. In contrast, for authentic student content, binary classification was comparatively more accurate (especially for programming), while likelihood estimates remained above 50\% but were generally lower than those for most LLM-generated responses, suggesting the detector may capture some signal even when imperfectly calibrated.

On the generation side, prompt variants intended to mimic students affected detectability primarily in the natural language tasks. Introducing typographic errors reduced detectability compared to baseline or persona prompts. For programming, the persona prompt had little effect relative to baseline, while inserting beginner bugs had an inconsistent impact across the two questions. Response length also differed substantially: despite using the same exercise descriptions given to students (including implicit guidance such as ``in a sentence'' versus ``several sentences to several paragraphs''), LLM outputs were markedly longer than student submissions. Length may therefore provide one cue the detector exploits, but it does not explain the short-answer reversal, where real student answers were judged more LLM-like.

\subsection{What Triggers Detection}
For RQ3, \textit{What characteristics of LLM-generated responses contribute to successful or failed detection?}, 
the observed patterns suggest that successful detection relies on persistent, domain-specific signatures of LLM output. In reflection tasks, unusually polished and verbose prose appears to increase detectability, and in programming tasks, regular, high-level structure may provide consistent cues. Detection failed most often when these signatures were weakened or masked, such as when LLM outputs were deliberately obfuscated to resemble novice writing (e.g., typos and informality), and when the task constrained responses to short, heterogeneous natural language (short answers), where authentic student responses could plausibly resemble templated or generic explanations.

\subsection{Limitations}
First, our detectability results reflect a single LLM-based detector and may not generalize to other detection models or non-LLM approaches. Second, the data comes from a single course context, so student response styles and exam formats may differ in other settings. Third, for programming tasks we removed non-code content from LLM outputs to match the student submission format; different preprocessing choices could affect detectability. Finally, prompting conditions differed by exercise type (typos for textual tasks; bugs for programming tasks), so prompt effects should be interpreted within each exercise type rather than compared directly across types.

\section{Conclusion}
Taken together, our findings indicate that LLM self-detection can be informative in some educational settings, particularly for programming and longer free-form writing, but it is unreliable for short-answer formats and sensitive to relatively small prompt changes. Consequently, LLM-based detection should not be treated as a standalone evidence source for academic integrity decisions, and caution is especially warranted in contexts dominated by short, formulaic textual responses.

\newpage
\section*{Acknowledgments}
This work was supported by Research Council of Finland grant \#356114. Generative AI was used for polishing the text in this paper. All GenAI-generated text was reviewed by the (human) authors and the human authors take full responsibility for the text.

%
\bibliographystyle{abbrv}
\bibliography{references}  
%

\balancecolumns
\end{document}